# A Random Forest-based Prediction Model for Turning Points in Antagonistic Event-Group Competitions

*Zishuo Zhu*

*School of Artificial Intelligence and Data Science*
*Hebei University of Technology*
*Tianjin 300401, China*

215492@stu.hebut.edu.cn

***Abstract* —At present, most of the prediction studies related to antagonistic event-group competitions focus on the prediction of competition results, and less on the prediction of the competition process, which can not provide real-time feedback of the athletes' state information in the actual competition, and thus can not analyze the changes of the competition situation. In order to solve this problem, this paper proposes a prediction model based on Random Forest for the turning point of the antagonistic event-group.**

**Firstly, the quantitative equation of competitive potential energy is proposed; Secondly, the quantitative value of competitive potential energy is obtained by using the dynamic combination of weights method, and the turning point of the competition situation of the antagonistic event-group is marked according to the quantitative time series graph; Finally, the random forest prediction model based on the optimisation of the KM-SMOTE algorithm and the grid search method is established. The experimental analysis shows that: The quantitative equation of competitive potential energy can effectively reflect the dynamic situation of the competition; The model can effectively predict the turning point of the competition situation of the antagonistic event-group, and the recall rate of the model in the test set is 86.13%; The model has certain significance for the future study of the competition situation of the antagonistic event-group.**

## I. Introduction

A group of sports with similar competitive characteristics and training requirements can be classified into a group of sports in which the process of ‘attacking’ and ‘defending’ and the strategy of ‘playing oneself’ and ‘limiting the opponent’ are referred to as an antagonistic event-group. The sports with the process of ‘attack’ and ‘defence’ and the strategy of ‘playing oneself’ and ‘limiting the opponent’ are called antagonistic sports. Competitive situations in rivalry sports are dynamic, with fluctuations in the performance of athletes due to a variety of factors, resulting in a turnaround of the competition. In the men's singles final at Wimbledon 2023, Carlos Alcaraz repeatedly turned the tables on adversity to win the match and end Novak Djokovic's Grand Slam run since 2013. Effectively predicting upcoming changes in competitive situations in rivalry events can help athletes to understand their own competitive status and to develop the right strategies to gain a competitive advantage.

The situation of the competition produces a turn, is the result of changes in the performance of athletes, to find the factors affecting the performance of athletes and quantify is the key and difficult to predict the situation of the competition. There are the following researches to measure the changes of athletes ‘performance in the competition, Zheng X. [1] called the change law of athletes’ competitive ability in time as ‘competitive rhythm’, and its influencing factors are the athletes ‘competitive ability of both sides, the evaluation behaviour of referees, the coaches’ on-field tactical command, pre-match preparations and trainings, spectators and venues, etc. Chen L [2] proposed that competitive potential energy is the internal factor affecting competitive performance, which consists of the athletes' competitive ability, competitive state, subjective power level when dealing with different situations in the competition, and the coach's on-field command ability and play level. Luo G. [3] study of the phenomenon of random ups and downs in collective sports shows that the critical point of ups and downs in competitive performance is both an opportunity and a danger, which can be judged by the changes in the structure of the personnel of the competition, the changes in the structure of the time, the different confrontation situations, the changes in the offensive tactics, and the changes in the local functions.

Currently, most of the related prediction researches focus on the result prediction [4-8], with less prediction on the amount of the process of the competition, thus less real-time feedback to the coaches and players in the actual competition. Liu et al [7] used six machine learning methods to establish a prediction model for women's speed skating performance, which was used to judge whether an athlete could participate in the 5000 competition or win a medal. Yang [8] constructed a functional equation model for predicting the winners and losers of tennis matches based on the technical statistics of tennis matches after the matches through the discriminant analysis method in mathematical statistics, which can effectively predict the winners and losers of men's tennis singles matches. Some studies also focus on the prediction of athletes' performance on the field of play. Manish et al. [9] used machine learning and deep learning to build a model for predicting the future performance of players in different tactical positions in football matches. However, none of the above literature can describe the competition between multiple players and cannot analyze the overall situation bias of the competition. Therefore, there is a need to develop a model for predicting turning points in the antagonistic event-group competitions, which not only reflects changes in the situation of the competition and predicts turning points, but also reflects real-time information about the performance of multiple players during the competition to help athletes and coaches to adjust their competition strategies.

In recent years, machine learning techniques have evolved and they are widely used in the field of sports competitions [10-12]. A review by Horvat et al [12] reviewed the application of machine learning in sports outcome prediction and pointed out that the existing prediction models are only applicable to a particular competition, and the development of one and the same model for predicting multiple competitions can be better for comparative analysis. The Random Forest algorithm is very tolerant to outliers and noise in machine learning and is not prone to overfitting [13], which is suitable for dealing with complex nonlinear relationships and high-dimensional data. Therefore, this paper will use the Random Forest algorithm to establish a prediction model of the turning point in the antagonistic event-group competitions, which is applicable to the same in the antagonistic event-group (e.g., tennis, badminton, and volleyball in the net separating antagonistic event-group), and the model is of some significance for the future study of the situation of the antagonistic event-group competitions.

This paper proposes a prediction model of the turning point of the competitive situation in the antagonistic event-group competition based on random forest, which first quantifies the competitive potential energy by establishing the relevant indicators of the competition and using the dynamic combination weight method. Secondly, the turning point of the competitive situation of the competition is marked by drawing the timing diagram of the competitive potential energy of both sides of the athletes. Since the turning point is a minority class sample relative to the non-turning point, the KM-SMOTE algorithm is used for oversampling, and then the random forest prediction model is trained. Finally, the experimental analysis verifies that the model has good accuracy.

## II. METHODS

The model constructed in this paper is mainly composed of two parts : the quantification of competitive potential energy, the optimization and training of random forest. The following will elaborate the steps and principles, and the process is shown in Fig.1.

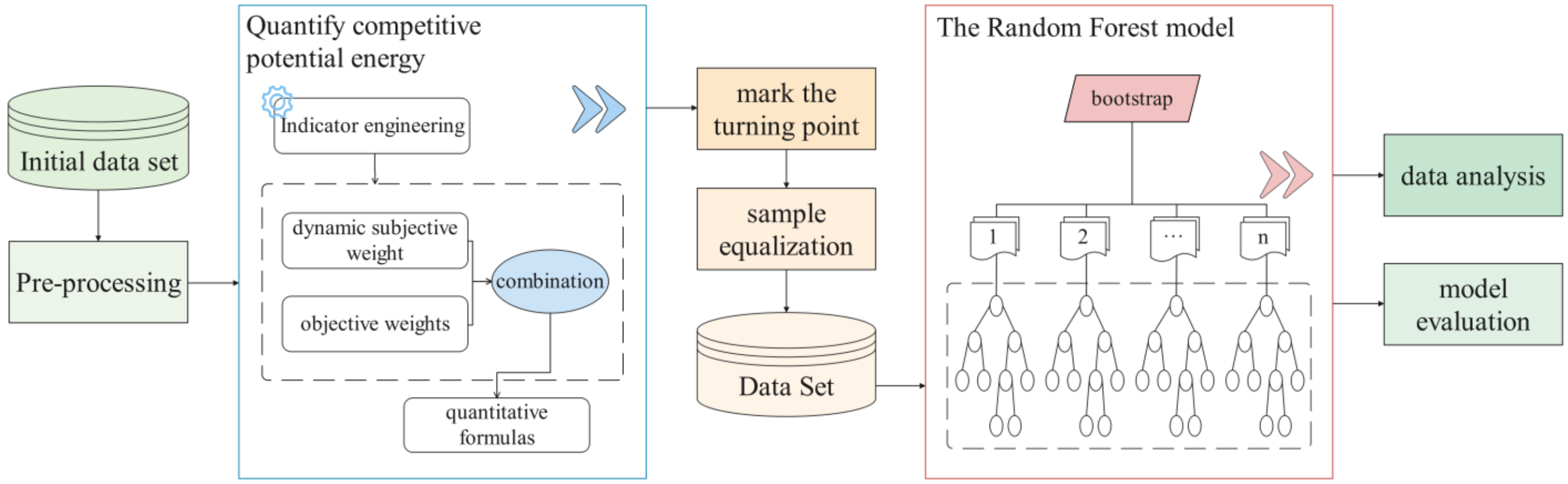

Fig. 1. The flow chart

### A. *Definition and quantification of the competitive potential energy*

Competitive potential energy is the driving force of the athletes in different competition time units ( the period from serve to score ), due to the change of their own and opponent 's competitive ability and the comprehensive influence of other factors. The change of competitive potential energy of athletes on both sides determines the trend of the competition situation. Similar to the definition of ' potential energy ' in physics, when an athlete 's competitive potential energy is high, his scoring emotion is high, and he is easier to control the competition. Competitive potential energy is a cumulative quantity, that is to say, the competitive potential energy of the previous competition time unit will affect the competitive potential energy of the current competition time unit. For example, the continuous score of the athlete in the previous competition time unit leads to a positive state of play in the current competition time unit.

The influencing factors of competitive potential energy include the athlete 's own competitive ability, the competitive ability of the opponent, the referee 's evaluation behavior, the coach 's on-the-spot tactical command, the audience and the venue. According to Equation (1), the competitive potential energy is quantified :

$$P = ( S + O + T ) \times W^T + P' \tag{1}$$

where $P$ is the quantitative value of competitive potential energy, $S$ is the individual performance factor, $O$ is the opponent 's performance factor, $T$ is the other influencing factors, $W$ is the weight vector of the three factors, and $P'$ is the potential energy of the previous competition time unit. $S$, $O$ and $T$ each include different indicators, such as factor $S$, $O$ factors including physical fitness, psychology, skills, tactics and other indicators, $T$ includes the referee 's assessment behavior, coach command, audience and venue and other indicators [1] . The indicators of different factors need to be determined according to different competition contents and real-time conditions. It should be noted that in order to eliminate the influence of different dimensions of each indicator, it is necessary to standardize the indicator data set to quantify the competitive potential energy.

### B. The establishment of dynamic combination weight

The weight established by a single method cannot take into account both subjectivity and objectivity, and the weight of each factor indicator in different stages of the antagonistic event-group competition is variable ( such as the decline of athletes ' physical strength in the later stage of the competition, and the influence of physical factors is significantly stronger than that in the early stage of the competition ), the use of a single method to obtain the weight is undoubtedly one-sided. Therefore, the method of combining subjective weight and objective weight is adopted, and the subjective weight is dynamically adjusted in each stage of the competition, and then reasonably combined with the objective weight, so as to obtain the indicator weight which is more scientific and close to the actual situation of the antagonistic event-group competition.

#### 1) The establishment of dynamic subjective weight

The subjective weight can be obtained by using the Analytic Hierarchy Process (AHP) [14], AHP is a quantitative analysis method proposed by Thomas Satti in the 1970 s to deal with multi-objective decision-making problems, which is mainly used to help decision-makers compare and weigh different factors in order to make the best decision. The steps of AHP are as follows : (1) Establish a hierarchical structure, including the target layer, the criterion layer and the scheme layer; (2) For the two levels, a pairwise judgment matrix is established, and the importance level labeling takes the number between 1 and 9 and its reciprocal; (3) Consistency test, only the consistency test through (consistency indicator CI < 0.1) judgment matrix can be accepted [15]; (4) The weight of each element is calculated by using the judgment matrix between each level.

#### 2) The establishment of objective weight

Entropy weight method is an objective weighting method, which can enhance the difference of indicators to avoid the influence of too small difference of selected indicators on the most results, and can effectively overcome the disadvantages of subjective factors [16]. The principle of entropy weight method is based on the degree of variation of indicators, and the higher the degree of variation, the higher the corresponding weight. Assuming that there are n evaluation indicators and m objects to be evaluated, the entropy of the ith evaluation indicator is calculated according to Equation (2):

$$H_j = -k \sum_{i=1}^{m} f_{ij} \ln f_{ij} \tag{2}$$

where $H_j$ is the entropy of the $j$th evaluation indicator; $k = \frac{1}{\ln m}$ ; $H_j \geq 0, k \geq 0$ ; when $f_{ij} = 0$ , $f_{ij} \ln f_{ij} = 0$ ; $f_{ij}$ is calculated according to Equation (3) :

$$f_{ij} = -\frac{Z_{ij}}{\sum_{i=1}^{m} Z_{ij}} \tag{3}$$

where $Z_{ij}$ is the standardized indicator value of the $i$ th evaluation object under the $j$th evaluation indicator.

The entropy weight $w_j$ of the $j$ th evaluation indicator is calculated according to Equation (4) :

$$w_j = \frac{1-H_j}{n-\sum_{j=1}^{n} H_j} \tag{4}$$

#### 3) The combination of weights

The subjective and objective weights are combined with the commonly used Equation (5):

$$W_i = \frac{\alpha_i \cdot \beta_i}{\sum_{i=1}^{n} \alpha_i \cdot \beta_i} \ , \ i = 1,2,\cdots,n \tag{5}$$

where $W_i$ represents the combined weight of the $i$ th indicator, $\alpha_i$ represents the subjective weight of the $i$th indicator, and $\beta_i$ represents the objective weight of the i th indicator.

When calculating the combined weight of the above equation, there are the following problems [17]: when the subjective weight is equal to the objective weight, the calculated combined weight deviates from the actual value; the combined weight results make the large weight become larger and the small weight become smaller; the deviation of the results calculated by the combination of individual weights is large. Therefore, the Equation (6) is used to combine the weights:

$$W_i = \frac{\alpha_i + \beta_i}{\sum_{i=1}^{n} (\alpha_i + \beta_i)} \ , \ i = 1,2,\cdots,n \tag{6}$$

where $W_i$ represents the combined weight of the $i$ th indicator, $\alpha_i$ represents the subjective weight of the $i$th indicator, and $\beta_i$ represents the objective weight of the i th indicator.

Note that because the subjective weight changes in different stages of the competition, the subjective and objective weights of different stages should also be combined separately.

### C. Mark the turning point of the competitive situation in the competition

By calculating the combination weight, the quantitative value of the competitive potential energy at different times of the competition can be obtained. The following will define the turning point of the competition situation.

The turning point of the competition situation is the result of the change of the competitive potential energy, so the turning point of the competition situation can be found by analyzing the change of the competitive potential energy of the athletes on both sides of the competition. As the name suggests, the turning point is the change of the competitive situation, in the antagonistic event-group competition, it is shown that the advantage of one side becomes the advantage of the other side or the advantage of the other side becomes the advantage of one side. Therefore, it is defined that the turning point of the competition situation is the intersection time of the time series curve of the competitive potential energy of the two sides of the athletes, that is, the competitive potential energy of the unit of the previous competition time is higher (lower) than that of the other side, and the competitive potential energy of the unit of the current competition time is lower (higher) than that of the other side. Then this is the turning point of the competition situation.

### D. Machine learning to predict the turning point of the competitive situation of the competition

Machine learning is divided into supervised learning and unsupervised learning. Supervised learning is used for data classification and prediction, and unsupervised learning is used for data clustering. The established competition situation turning prediction model is realized through supervised machine learning. The data set that needs machine learning training includes n indicators and response Y of S, O, T three factors, where Y is a 0-1 binary variable, 0 represents not a competition situation turning point, 1 represents a competition situation turning point.

In the antagonistic event-group competition, the turning point of the competition situation is a few moments relative to the ordinary time point, so the data set is an unbalanced sample set. The results of machine learning training for unbalanced data sets are more accurate for majority class samples, however, we focus on the prediction results of minority class samples. There are two methods to solve this problem, one is undersampling of majority class samples, and the other is oversampling of minority class samples [18]. Both methods can make the two types of samples in a more balanced state.

*1) RUSBOOST algorithm*

RUSBoost combines undersampling with AdaBoost algorithm [19]. This method first extracts a certain amount of majority class samples and minority class samples from the data set to form a training set, and then adjusts the penalty for misclassification each iteration to construct a base learner. Finally, the results are predicted by weighted voting of the base learner. The RUSBoost algorithm process is shown in Table 1:

TABLE I. THE RUSBOOST ALOGRITHM

| |
|---|
| **Algorithm 1:** RUSBoost |
| **Input:** Data set $D = \{(x_1, y_1), (x_2, y_2), \cdots, (x_m, y_m)\}$; Base learning algorithm h; The number of cycles of learning T. |
| **OutPut:** $H(x) = sign\left(\sum_{t=1}^{T} \alpha_t h_t(x)\right)$ |
| **Begin** |
| 1. $w_1(i) = \frac{1}{m}$ |
| 2. $\boldsymbol{for}\ t = 1,2,\cdots,T$: |
| 3. Under-sampling using the distribution $w_t$ |
| 4. Create the temporary data set $D_t$ |
| 5. $\varepsilon_t = E_{X-w_t}[y \neq h_t]$ |
| 6. ***if*** $\varepsilon_t > 0.5$ then |
| 7. jump out of the loop |
| 8. ***endif*** |
| 9. $\alpha_t = \frac{1}{2}\ln\left(\frac{1-\varepsilon_t}{\varepsilon_t}\right)$ |
| 10. $w_{t+1}(i) = \frac{w_t(i)e^{-\alpha_t y_i h_t(x_i)}}{z_t} = \frac{w_t(i)}{z_t}\begin{cases}\exp(-\alpha_t) & h_t = y_i \\ \exp(\alpha_t) & h_t \neq y_i\end{cases}$ |
| 11. ***endfor*** |
| **end** |

where $m$ is the number of samples, $w$ is the instance distribution, $\alpha_t$ is the weight corresponding to the prediction $h_t$, and $H(x)$ is the decision result of the learner.

*2) KM-SMOTE algorithm*

Chawla et al. [20] proposed the SMOTE algorithm. SMOTE (Synthetic Minority Over-sampling Technique) is a synthetic sampling technique used to deal with class imbalance problems. It balances the class distribution by linear interpolation in the feature space to synthesize new minority class samples, thereby improving the performance of the classifier on minority samples. Suppose the oversampling multiple is N, search K adjacent samples of each minority class sample and randomly select N samples, and interpolate between the original sample data and its adjacent samples. The new sample is synthesized by the Equation (7):

$$x_{new} = x + rand(0,1) \times (\theta_i - x) \quad (7)$$

where $x_{new}$ is a new sample of composition, $x$ is a minority class sample, $rand(0,1)$ is a random number between 0 and 1, $\theta_i$ is the $i$th adjacent sample of $x$, $i = 1,2,\cdots,N$.

The SMOTE algorithm effectively improves the number of minority class samples, but its algorithm has the following problems [21]: vulnerable to noise samples, new samples will most likely fall in the majority class area and become noise ; when synthesizing minority class samples, the distribution of majority samples is not considered, and new samples may fall in the overlapping area of the two types of samples. It is not easy to identify a small number of samples in sparse areas. Chen Bin et al. proposed KM-SMOTE algorithm [22] by combining K-means algorithm with SMOTE algorithm, which effectively avoids the change of rare class data distribution caused by interpolation data and solves the problem of fuzzy positive and negative class edges. The algorithm flow is as follows:

Step1: Determine the initial cluster $K$ value;

Step2: The *K-means* algorithm is used to divide the minority class samples into K clusters, and the cluster center of each cluster is $\{c_1, c_2, c_3, \cdots, c_K\}$;

Step3: Sample interpolation is performed on each cluster center, and the Equation (8) is synthesized:

$$x_{new} = c_i + rand(0,1) \times (X - c_i) \quad (8)$$

where $x_{new}$ is the new sample, $c_i$ is the cluster center, $rand(0,1)$ is a random number between 0 and 1, $X$ is the original sample data with $c_i$ as the cluster center, $i = 1,2,\cdots,K$.

*3) The Random Forest model*

Random Forests (RF) is a classical machine learning algorithm, which generates many decision trees by bootstrap random sampling. Multiple decision trees are integrated to form a random forest, and multiple weak learners are constructed into a strong learner. Its modeling principle is shown in Fig. 2. Each subsample is randomly sampled from the original data for decision tree modeling, and the results of the decision tree are combined to determine the results of the random forest.

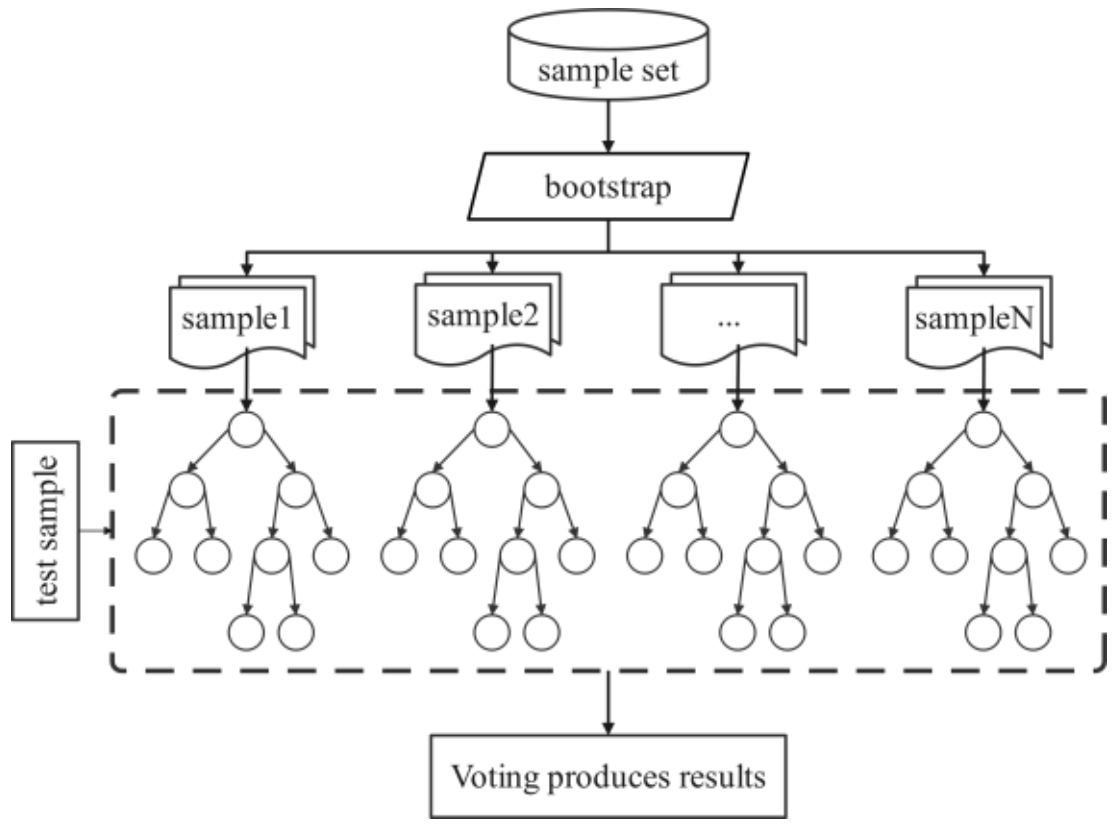

Fig. 2. The RF model

An important role of random forest is to measure the importance of indicators. In the random forest, after the sampling is put back, the remaining data becomes out-of-bag (OOB data). Through noise interference, the value of a feature vector in the OOB is changed, and the accuracy of the original OOB is subtracted from the accuracy of the OOB after adding noise, and the importance measure of this feature vector can be obtained. By analyzing the importance of indicators, coaches can help coaches better guide athletes to adjust their competition status.

## III. EXPERIMENT

### A. Data from the experiment

The research data are from the C problem in the MCM of the 2024 American College Students ' Mathematical Competition in Modeling. The content is the 2023 Wimbledon men 's singles competition numbered 1301-1701, a total of 31 competitions. The data cover a total of 7284 data such as competition system, score and technical action. A total of 14 indicators are extracted as shown in Table 2:

TABLE II. THE INDICATORS

| Indicator | Meaning | Type |
|---|---|---|
| X1 | Running distance (A) | Negative |
| X2 | Number of consecutive scores (A) | Positive |
| X3 | Number of consecutive errors (A) | Negative |
| X4 | Number of ace balls (A) | Positive |
| X5 | Number of winning balls (A) | Positive |
| X6 | Number of effective near-net (A) | Positive |
| X7 | Number of effective service breaks (A) | Positive |
| X8 | Running distance (B) | Positive |
| X9 | Number of consecutive scores (B) | Negative |
| X10 | Number of consecutive errors (B) | Positive |
| X11 | Number of ace balls (B) | Negative |
| X12 | Number of winning balls (B) | Negative |
| X13 | Number of effective near-net (B) | Negative |
| X14 | Number of effective service breaks (B) | Negative |

Among them, A represents the athlete himself and B represents the athlete's opponent. Because the indicator information of factors such as referees, spectators and venues is not provided and difficult to measure, only the competitive ability of athletes themselves and competition opponents is used as a measure of competitive potential energy. Competitive ability is measured from four aspects: physical fitness, psychology, skills and tactics : physical fitness is expressed by the distance of athletes in a scoring run; psychologically, consider continuous scores and continuous errors, including double errors and unforced errors; in terms of skills, through the description of the professional and technical ability of athletes, including ACE ball, winning ball and effective breaking three indicators; the tactical aspect is described by the tactical effectiveness adopted by the athlete, considering whether the athlete is effective near the net.

### B. Indicators for model evaluation

The turning point of the competition situation in the competition is a minority class sample relative to the non-turning point. It is unreasonable to use the accuracy of classification as an evaluation indicator only for the data set with unbalanced samples, because when there are few samples in the minority class, the classifier only needs to classify all samples into the majority class to achieve high accuracy, and it is concerned with the accuracy of minority class prediction. Therefore, this paper uses classification accuracy, recall rate and G-mean as the evaluation indicators of the model.

The majority class sample is defined as a negative class, and the minority class sample is defined as a positive class. The following four phenomena will occur in the classification as shown in Table 3:

TABLE III. THE CONFUSION MATRIX

| Classification | Actual P class | Actual N class |
|---|---|---|
| **Classified as P class** | TP | FP |
| **Classified as N class** | FN | TN |

where TP represents the sample that is actually a positive class is classified as a positive class; FP represents that the samples that are actually negative are classified as positive; FN represents the classification of samples that are actually positive as negative; TN represents the actual negative class of samples classified as negative class.

The classification accuracy represents the ratio of the number of correctly classified samples to the total number of samples. It can reflect the overall performance of the model and is calculated according to Equation (9):

$$Accuracy = \frac{TP+TN}{TP+FN+FP+TN} \times 100\% \tag{9}$$

The Recall represents the percentage of samples that are actually positive that are correctly predicted, and it can reflect the accuracy of the classification of the minority class samples that we are more concerned about. Calculating by Equation (10):

$$Recall = \frac{TP}{TP+FN} \times 100\% \tag{10}$$

G-mean represents several average values of the classification accuracy of minority class samples and the classification accuracy of majority class samples. When the classification accuracy of both categories is high, the value of G-mean is the largest. Calculated according to Equation (11):

$$G\text{-}mean = \sqrt{\frac{TP}{TP+FN} \times \frac{TN}{FP+TN}} \times 100\% \tag{11}$$

### C. Steps of the experiment

According to the method of the first section, the experiment is carried out: first, the competitive potential energy should be quantified; then, write the algorithm code to mark the competitive turning point of the competition to form the original data set; the original data set is processed by KM-SMOTE algorithm, and the random forest model is established and adjusted.

#### 1) Quantify competitive potential energy and mark the turning point of competition.

Here, the time unit is from the beginning of a serve to the end of a score in a tennis match. The competitive potential energy indicators of both players in each time unit in a match are counted and quantified, and then analyzed.

According to the Equation (1), the standardized indicators are linearly added to quantify the competitive potential energy, and the very large indicator symbol is the negative sign of the small indicator symbol, and the Equation (12) is obtained.

$$P = (-X_1 + X_2 - X_3 + X_4 + X_5 + X_6 + X_7 + X_8 - X_9 + X_{10} - X_{11} - X_{12} - X_{13} - X_{14}) \times W^T + P' \quad (12)$$

where $P$ is the quantitative value of competitive potential energy, $X_{1-14}$ is the 14 indicators of competitive potential energy, $W$ is the weight vector of 14 indicators, and $P'$ is the potential energy of the previous competition time unit.

When obtaining the subjective weight, the competition is divided into two stages: the first half (a) and the second half (b). The AHP method is used to obtain the subjective weight of the two stages respectively, so that the influence of time change on the indicator weight can be reasonably described. Taking the competition 2023-wimbledon-1301 Carlos Alcaraz against Nicolas Jarry as an example, according to the model constructed above, the weight of each indicator under this competition is obtained as Table 4:

TABLE IV. COMBINATION WEIGHT

| Objective | First-level indicator | Second-level indicator | Weight | | | | |
|---|---|---|---|---|---|---|---|
| | | | AHP (a) | AHP (b) | Entropy | Combination (a) | Combination (b) |
| Quantify competitive potential energy | Athletes ' own competitive ability | X1 | 0. 0496 | 0. 2403 | 0. 0113 | 0. 0305 | 0. 1258 |
| | | X2 | 0. 1675 | 0. 1254 | 0. 0264 | 0. 0970 | 0. 0759 |
| | | X3 | 0. 1675 | 0. 1254 | 0. 0757 | 0. 1216 | 0. 1006 |
| | | X4 | 0. 0837 | 0. 0594 | 0. 1041 | 0. 0939 | 0. 0818 |
| | | X5 | 0. 0837 | 0. 0594 | 0. 0644 | 0. 0741 | 0. 0619 |
| | | X6 | 0. 0573 | 0. 0284 | 0. 0893 | 0. 0733 | 0. 0588 |
| | | X7 | 0. 0573 | 0. 0284 | 0. 1325 | 0. 0949 | 0. 0804 |
| | Athletes ' competitive ability of opponents | X8 | 0. 0248 | 0. 1201 | 0. 0112 | 0. 0180 | 0. 0657 |
| | | X9 | 0. 0838 | 0. 0627 | 0. 0297 | 0. 0567 | 0. 0462 |
| | | X10 | 0. 0838 | 0. 0627 | 0. 0616 | 0. 0727 | 0. 0621 |
| | | X11 | 0. 0419 | 0. 0297 | 0. 0969 | 0. 0694 | 0. 0633 |
| | | X12 | 0. 0419 | 0. 0297 | 0. 0593 | 0. 0506 | 0. 0445 |
| | | X13 | 0. 0286 | 0. 0142 | 0. 0756 | 0. 0521 | 0. 0449 |
| | | X14 | 0. 0286 | 0. 0142 | 0. 1621 | 0. 0954 | 0. 0882 |

The obtained combination weight vector is brought into Equation (11), and the time series diagram of the competitive potential energy of both sides of the athletes can be obtained the time series diagram of the competitive potential energy of both sides of the athletes is shown in Fig. 3. The intersection of the competitive potential energy curves of both sides is the turning point of the competition situation. In this experiment, the explanation of the turning point in the indicator set is that there is a certain time unit of competition, when the positive indicator of the dominant side is small and the negative indicator is large, and when the positive indicator of the inferior side is large and the negative indicator is small.

Fig. 3 shows the fluctuation of the two 's competitive potential energy, and draws the winning situation of each competition from the post-match data (see Fig. 4). Combined with the analysis of Fig. 3 and Fig. 4: the first competition (time unit 1-63 points), the two alternately won the competition, and the last two competitions Carlos Alcaraz won the two competitions in a row by 5 : 3. From the competitive potential energy diagram, it can also be seen that the two in the time unit 0-45 points, the competitive potential energy is not much different, the time unit 45-63 points, Carlos Alcaraz competitive potential energy is significantly higher than Nicolas Jarry; the second set of competitions (time unit 64-155 points) began, Nicolas Jarry won three competitions in a row (time unit 64-87 points) to open the gap, Carlos Alcara recovered the state in the last few competitions to catch up with the score, but Nicolas Jarry still won the second set with 7 : 6.From the competitive potential energy diagram, it can also be seen that the advantages established in the early stage of Nicolas Jarry made his competitive potential energy rise, and Nicolas Jarry made efforts in the later stage but failed to change the outcome.

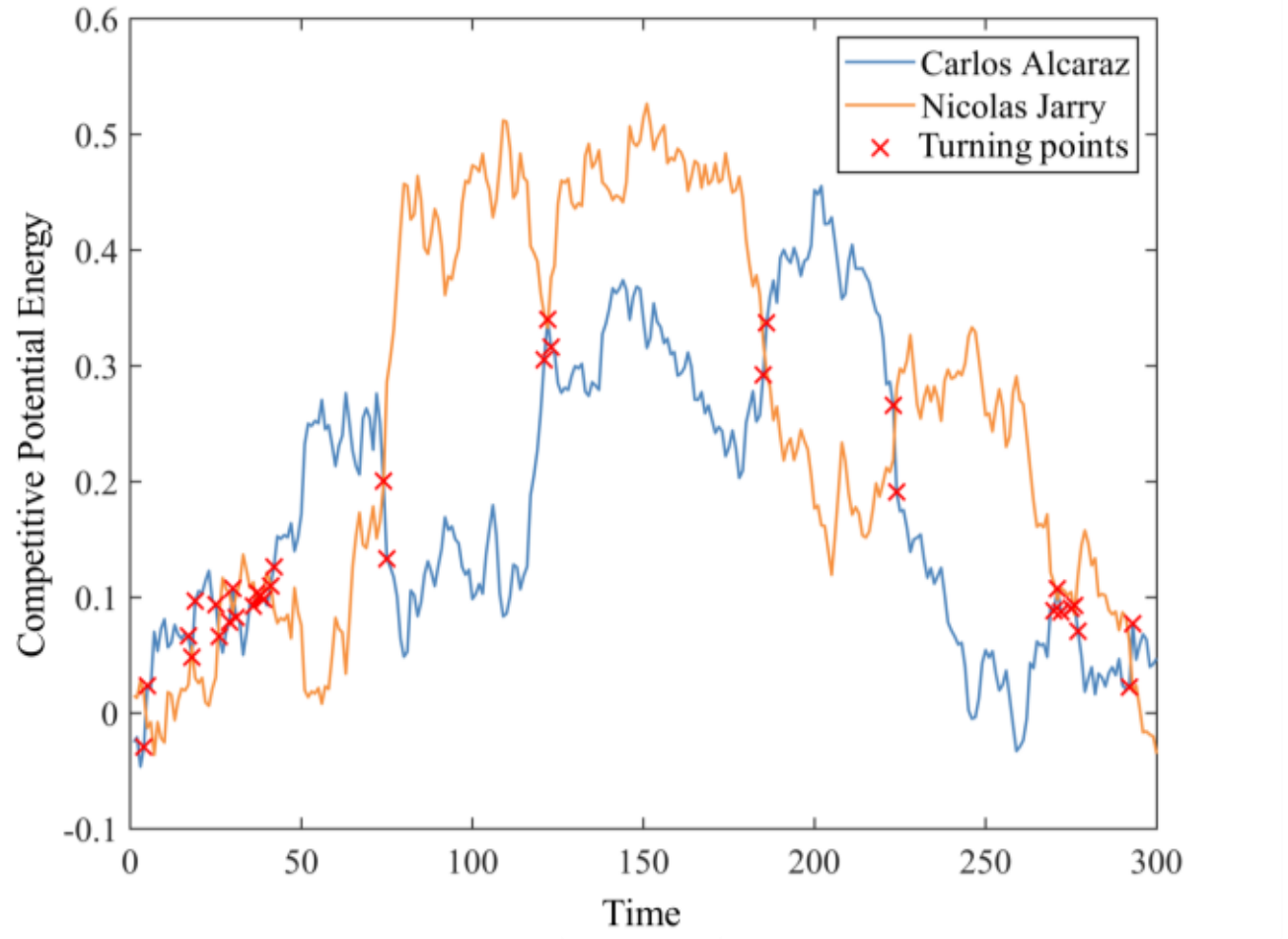

Fig. 3. Time series diagram of competitive potential energy

### 2) *Training the Random Forest model*

The first 80% of the data set is taken as the training set, and the last 20% is taken as the prediction set. MATLAB programming is used to realize KM-SMOTE oversampling of minority class samples, so that the final minority class samples are balanced with the majority class samples. The sample balance before and after oversampling is shown in Table 5:

TABLE V. SAMPLE EQUILIBRIUM STATE

| The original set | | | | After oversampling | | |
|---|---|---|---|---|---|---|
| Number of samples | Number of minority class samples | Number of majority class samples | Imbalanced ratio | Number of samples | Number of minority class samples | Number of majority class samples |
| 5827 | 532 | 5295 | 10. 04% | 10592 | 5297 | 5295 |

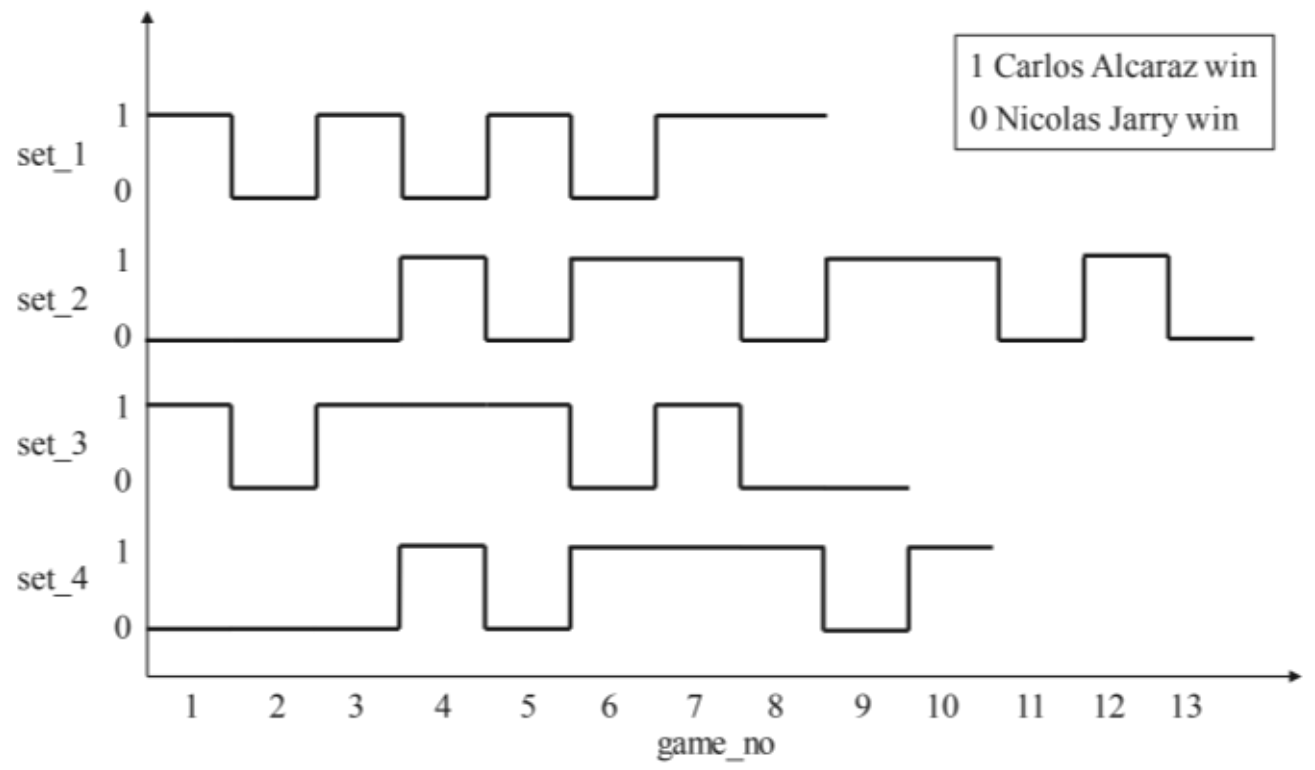

Fig. 4. The winner of each game of match 1301

Taking the maximum AUC as the target, the number of grid search decision trees and the maximum number of splits of the decision tree are two hyperparameters, and the grid search range is gradually reduced to find the optimal hyperparameter. Finally, the number of decision trees is 445, the maximum number of splits is 915, and AUC = 0.926.

### *D. Results and Analysis*

The super-parameters optimized by the grid method are brought into the model, and the 5-fold cross-validation is performed. The evaluation indicator is calculated by averaging the results of the classifier after 5 times of modeling. This paper uses MATLAB R2021 b to implement KM-SMOTE, grid search method, and random forest algorithm, where the KM-SMOTE algorithm has a cluster number of 3, an oversampling percentage of 90, and a proximity value of 5. The model training results are shown in Table 6 and Fig. 5.

TABLE VI. MODEL TRAINING RESULTS

| **Algorithm** | **Accuracy** | **Recall** | **G-mean** |
|---|---|---|---|
| KM-SMOTE+RF | 85. 23% | 88. 26% | 85. 18% |

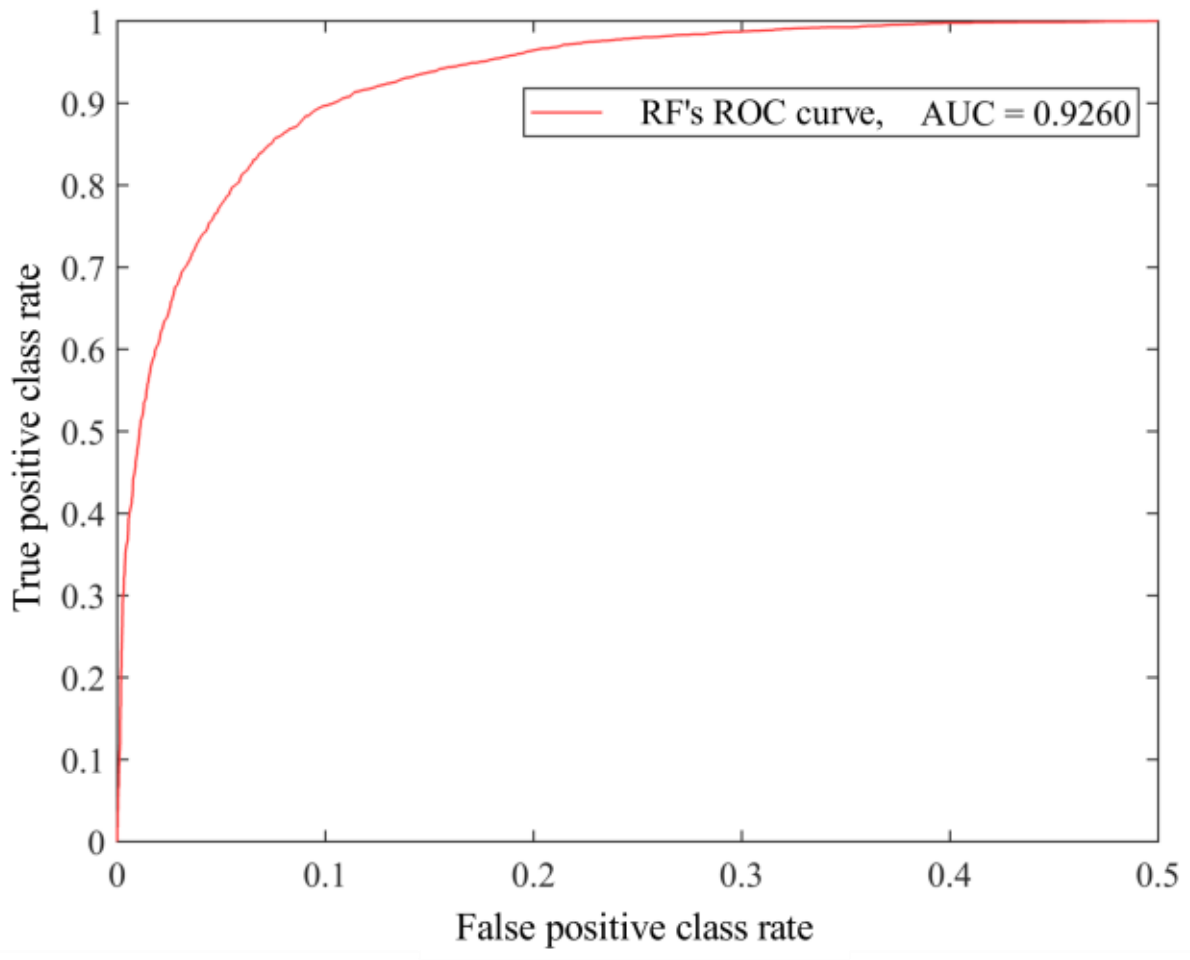

Fig. 5. Training set ROC curve

The classification accuracy, recall rate and G-mean of the model after training are all greater than 0.85, and the ROC curve is drawn and the calculated AUC value is greater than 0.9, indicating that the model has good performance on the training set.

The model is applied to the test set, and the performance of the model on the test set is tested to evaluate the generalization ability of the model. The results are shown in Table 7 and Fig. 6:

TABLE VII. MODEL TEST RESULTS

| **Date** | **Accuracy** | **Recall** | **G-mean** |
|---|---|---|---|
| Test set samples | 65. 48% | 86. 13% | 73. 86% |

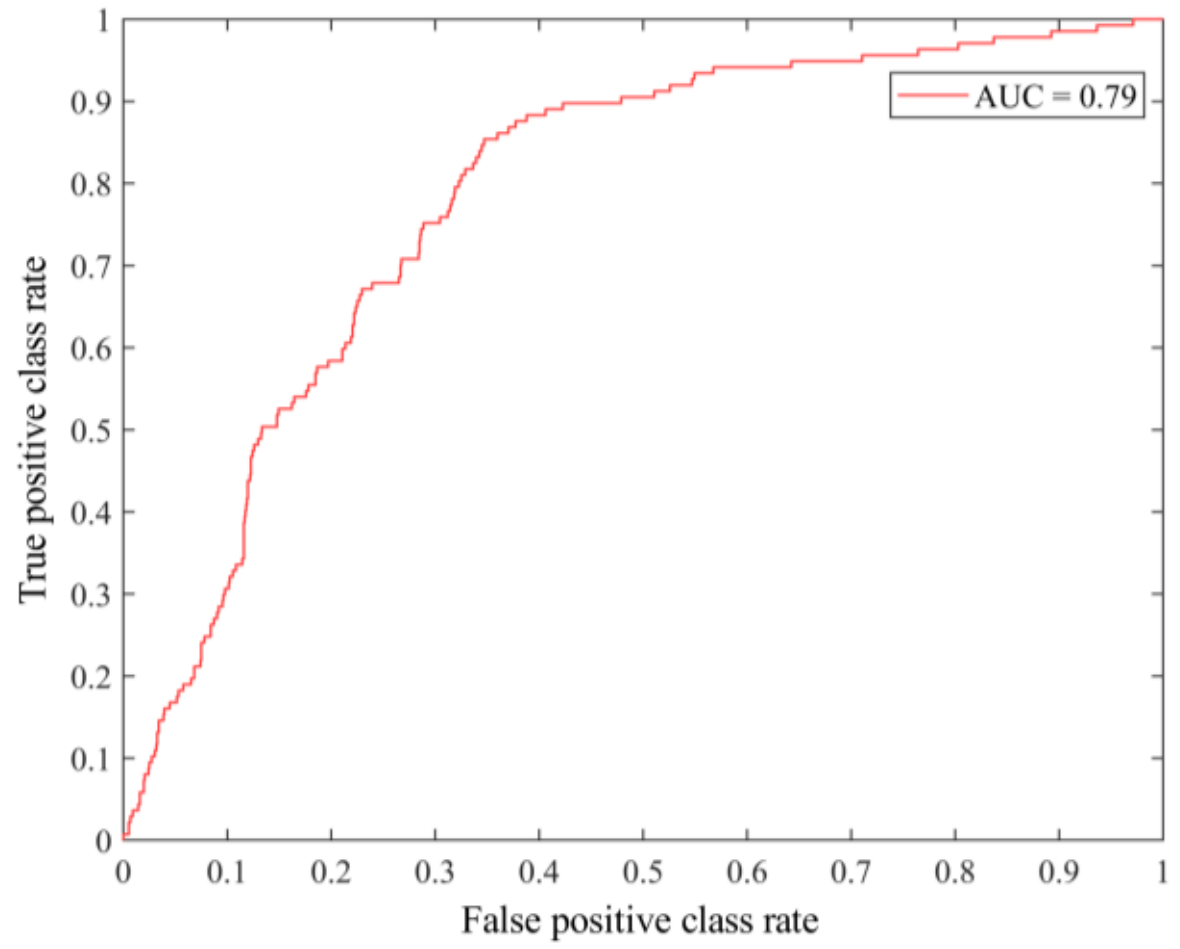

Fig. 6. Test set ROC curve

The recall rate of the model on the test set is greater than 85 %, indicating that the model can effectively predict the turning point of the vast majority of the competition situation in the unknown data set, but the prediction accuracy of the non-turning point is low, and the G-mean is 73.86 %. In general, the model has a certain ability to predict, draw its ROC curve and calculate the AUC is 0.79, indicating that the model has a good performance on the training set and has a certain generalization ability.

The importance of each indicator obtained by applying noise to the OOB data of the random forest model is shown in Fig. 7, among which the indicators with larger weights are X1 (own running distance), X5 (own number of winning shots), X8 (opponent running distance), X9 (opponent continuous scoring times), X10 (opponent continuous error times). Based on the importance of the indicators obtained by the model established in this experiment, it can be analyzed that X1, X5, X8, X9, X10 should be focused on in the competition. These indicators, when the relevant positive indicators of the dominant party are smaller and the negative indicators are larger. At the same time, the relevant positive indicators of the inferior side are larger and the negative indicators are smaller, then there may be a turning point in the competition situation.

## IV. CONCLUSION REMARKS

This paper proposes a prediction model based on random forest for the turning point in the antagonistic event-group competition situation, which is used to predict whether the competitive situation has a turning point in a certain competition time unit. At the same time, it can also analyze the competitive state of the athletes according to the time series changes of the competitive potential energy of both sides of the athletes, so as to formulate the winning strategy. The innovations of the model are as follows: (1) This paper is devoted to establishing a universal model of all antagonistic event-group, which can be better compared and analyzed. (2) The model focuses on the amount of competition process. The model is not used to predict the results of the competition, but reflects the time series changes of the competitive potential energy of both sides during the competition, and can feedback the competitive state of the athletes of both sides in real time. (3) The weighting of indicators is innovative. As the competition changes dynamically over time, the weights of the indicators will also change, and the combination of dynamic AHP and entropy weighting method is used to assign the weights, taking into account the subjectivity and objectivity. Further research in the future will improve the competitive potential equation, consider how to measure off-field factors and incorporate them into the competitive potential equation. Other advanced machine learning strategies will also be considered.

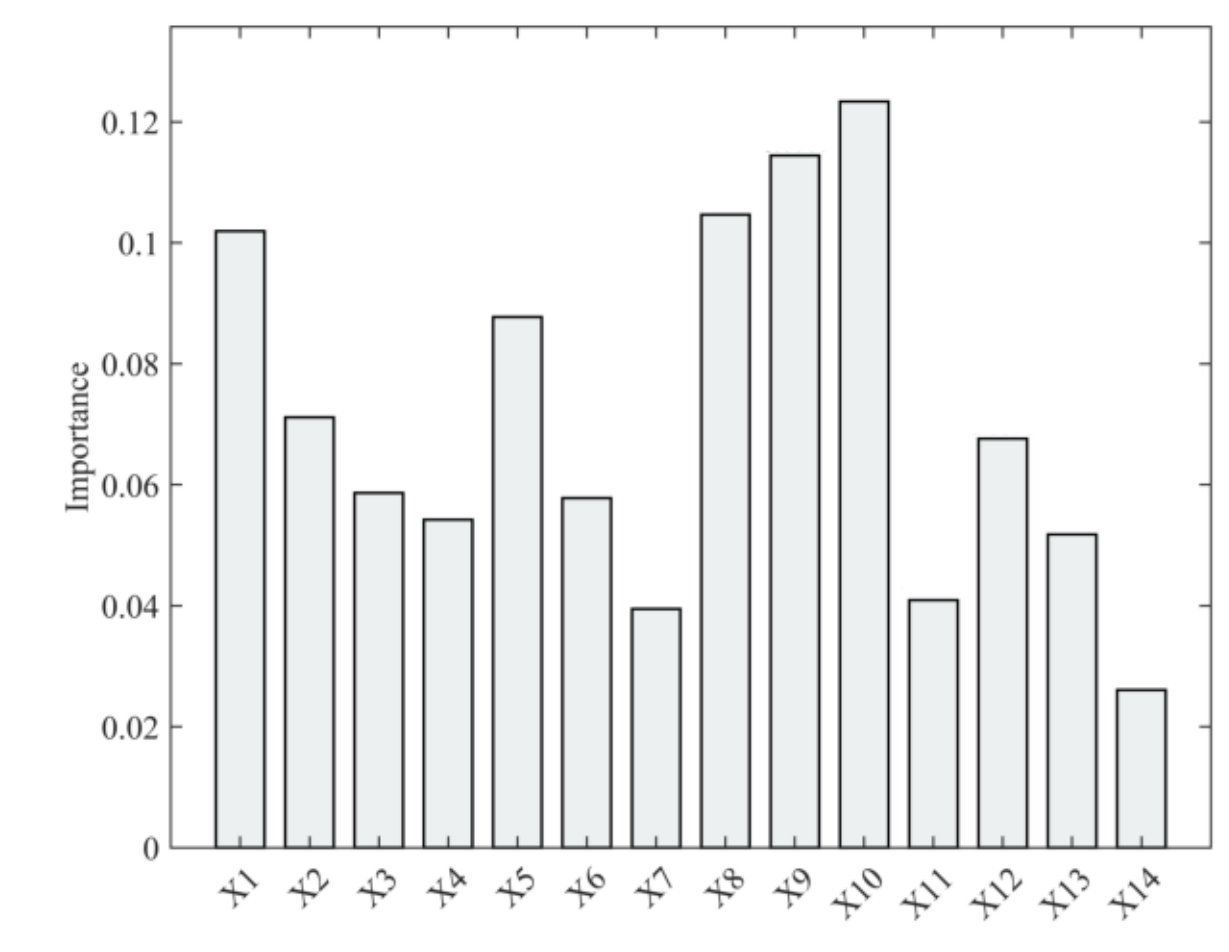

Fig. 7. Importance of indicators